\documentclass[sigconf]{acmart}

\usepackage{latexsym}
\usepackage{makecell}
\usepackage[english]{babel}

\AtBeginDocument{%
  \providecommand\BibTeX{{%
    \normalfont B\kern-0.5em{\scshape i\kern-0.25em b}\kern-0.8em\TeX}}}

\setcopyright{acmcopyright}
\copyrightyear{2021}
\acmYear{2021}
\setcopyright{iw3c2w3}
\acmConference[WWW '21]{Proceedings of the Web Conference 2021}{April 19--23, 2021}{Ljubljana, Slovenia}
\acmBooktitle{Proceedings of the Web Conference 2021 (WWW '21), April 19--23, 2021, Ljubljana, Slovenia}
\acmPrice{}
\acmDOI{10.1145/3442381.3449876}
\acmISBN{978-1-4503-8312-7/21/04}

\begin{document}

\title{Diverse and Specific Clarification Question Generation with Keywords}


\author{Zhiling Zhang}
\affiliation{%
  \institution{Shanghai Jiao Tong University}
  \city{Shanghai}
  \country{China}}
\email{blmoistawinde@sjtu.edu.cn}
\orcid{0000-0002-8081-704X}

\author{Kenny Q. Zhu}
\affiliation{%
  \institution{Shanghai Jiao Tong University}
  \city{Shanghai}
  \country{China}}
\email{kzhu@cs.sjtu.edu.cn}
\authornote{Corresponding author.}

\begin{abstract}
  Product descriptions on e-commerce websites often suffer from 
missing important aspects. \textit{Clarification question generation} (CQGen) can be a promising approach to help alleviate the problem. Unlike traditional QGen assuming the existence of answers in the context and generating questions accordingly, CQGen mimics user behaviors of asking for unstated information. The generated CQs can serve as a sanity check or proofreading to help e-commerce 
merchant to identify potential missing information before advertising their
product, and improve consumer experience consequently. 
Due to the variety of possible user backgrounds and use cases, 
the information need can be quite diverse but also specific to a detailed
topic, while previous works assume generating one CQ per context and 
the results tend to be generic. We thus propose the task 
of \textit{Diverse CQGen} and also tackle the challenge of specificity. 
We propose a new model named \textit{KPCNet}, which generates CQs with 
Keyword Prediction and Conditioning, to deal with the tasks. 
Automatic and human evaluation on 2 datasets (\texttt{Home \& Kitchen}, 
\texttt{Office}) showed that KPCNet can generate more specific questions 
and promote better group-level diversity than several competing baselines. \footnote{Our code is available at \url{https://github.com/blmoistawinde/KPCNet}}
\end{abstract}


\maketitle

\section{Introduction}
\label{sec:intro}

\begin{figure}[htbp]
\centering
\includegraphics[width=\linewidth]{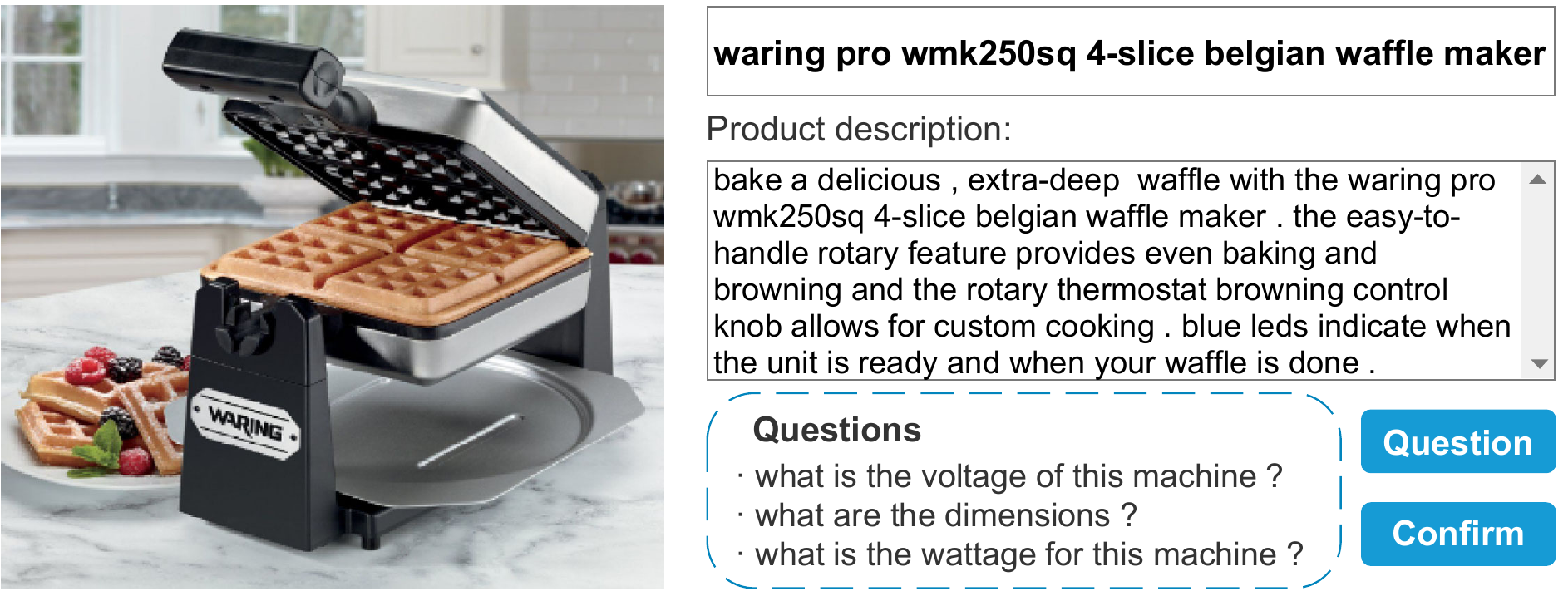}
\caption{A hypothetical writing assistant generating CQs.}
\label{fig:WA_UI}
\end{figure}
  
The development of the Internet has spawned a number of task-oriented writings, such as product descriptions on Amazon. However, since merchants cannot always have a thorough understanding of consumers' need due to the variety of possible user backgrounds and use cases, their writings usually miss something deemed important by the customers. For example, a US merchant may assume his device be 
used on a 110V power line, and thus omit this in the product description. 
Customers from Asia and Europe, where 220V is used, might pay special attention to the voltage 
requirements in the description. On finding this absent from the description, 
some customers may ask CQs like ``What is the voltage of this machine?'' in customer QA, while others would turn to other products immediately, an unfortunate loss to the seller. It would be helpful if the platform can provide a service to remind the merchants of those potentially missing needs with its broader knowledge.

Clarification question\footnote{questions asking for what's missing from a given context} generation (CQGen), which mimics the user engagement by raising questions, can be a promising option for such a service. Before publishing their writings, authors may request for CQs from somewhere like the hypothetical writing assistant we illustrated in Figure \ref{fig:WA_UI}, and supplement missing information accordingly. CQGen is a challenging task for the following reasons: 

First, it requires the question to be 
\textit{specific} while not being \textit{repetitive} to existing context. 
Questions pertaining to smaller set of products are considered more specific. 
For example, the first question in Figure \ref{fig:WA_UI} is more specific 
than the second one because it applies to only electric appliances, while
the second one applies almost to every product. 
In contrast to the traditional QGen task which is typically evaluated
on the SQuAD 1.1 dataset \citep{rajpurkar2016squad} and 
derives the specificity from the knowledge of answer, 
CQGen doesn't expect the existence of answer in the context. 
Therefore, QGen algorithms which require the answer span and its position
as input~\citep{song2018leveraging, sun2018answer,subramanian2018neural} 
do not apply here. Vanilla seq2seq model has been shown to generate 
highly generic questions by \citet{rao2019answer}. 
They then proposed GAN-Utility, which estimates the utility of answer with GAN 
as reward for RL to improve generation specificity. 
However, the answer used in the estimation is generated from the context and an already-generated 
question with another trained QA component, 
which may not be reliable here as the answers are inherently 
missing from context by definition. 
Consequently, this answer-based approach was shown to yield even worse results under 
some conditions~\citep{cao2019controlling}. We thus totally eliminate the need for answers in our work, 
which has the benefit of making use of more training data without answer.

Moreover, previous works on CQGen all assume generating one question per context. We claim that generating a group of diverse CQs 
(as is shown in Figure \ref{fig:WA_UI}) can be more beneficial, 
because this allows the system to efficiently cover a variety of user 
needs at once, and tolerate occasional errors as the rest questions are still useful. 
We name this novel task as \textbf{Diverse CQGen}. We seek algorithms that can deal with the task, and adopt a new group-level evaluation protocol to properly evaluate the effectiveness of algorithms under this scenario.

To deal with the specificity challenge, we propose a novel model named 
Keyword Prediction and Conditioning Network (KPCNet). Keywords in CQs is one kind of prior knowledge that the platform can mine about user needs. They are usually
product attributes or closely related concepts that make the questions specific, and thus the main semantic of a question can be captured by its keywords. 
For example, the keyword of ``What's the \textit{dimension}?'' 
is ``\textit{dimension}'', and the question can be comprehended even with 
a single word (``\textit{dimension}?''). We can generate more detailed question like ``Can you cook \textit{rice} in this \textit{cooker}?'' with keywords ``\textit{cooker, rice}''. 
Therefore, the proposed KPCNet first predicts 
the probability for a keyword to appear in the generated CQ, 
then selects keywords from the predicted distribution, 
and finally conditions on them to generate questions. 
We can also partially control the generation by operating on the 
conditioned keywords, which can be utilized to avoid repetitive questions 
and further improve the quality.

To promote diversity, we explore several diverse generation approaches for 
this problem, including model-based 
\textit{Mixture of Experts}~\citep{shen2019mixture} and 
decoding-based \textit{Diverse Beam Search} \citep{vijayakumar2018diverse}. 

KPCNet's controllability through keywords, on the other hand, 
enables keywords-based approaches. We explore a novel use of classic
clustering method on producing coherent keyword groups for keyword selection 
to generate correct, specific and diverse questions.

Individual and group-level evaluation showed that KPCNet is capable of 
producing more diverse and specific questions than strong baselines. 
Our contributions are:

\begin{enumerate}
  \item To our best knowledge, we are the first to propose the 
task of \textit{Diverse CQGen}, which requires generating a group of 
diverse CQs per context, to cover a wide range of information needs.  
(\S \ref{sec:intro})
  \item We propose KPCNet, which first predicts keywords that focus on the specific aspects of the question, 
before generating the whole question to promote specificity. (\S \ref{sec:specific}, \S \ref{sec:prediction})

  \item Based on KPCNet's keyword conditioned generation, we propose keyword selection methods to produce multiple keywords groups for generation diversity. (\S \ref{sec:diversity}, \S \ref{sec:cond_gen}, \S \ref{sec:selection})
  \item We show with probing tests that KPCNet can be further enhanced with external knowledge to alleviate the problem of asking existing information in the context, an under-explored yet fundamental problem in CQGen, and improve generation quality. (\S \ref{sec:probing})
\end{enumerate}

\section{Preliminaries}

\subsection{Keyword-based Diverse CQGen}

Given a textual \textit{context} $\mathbf{x} = (x_1, x_2, ..., x_{T_1})$, our aim is to generate a \textit{clarification question} $\mathbf{y} = (y_1, y_2, ..., y_{T_2})$, so that $y$ asks for relevant but not repetitive information to $\mathbf{x}$. In the setting of \textit{Diverse CQGen}, we should generate a group of CQs for the same context such that they are semantically different from each other. In this work, we additionally consider \textit{keywords} $\mathbf{z} = (z_1, z_2, ..., z_k)$ that are expected to capture the main semantic of $\mathbf{y}$. The definition of keywords may vary across domains, and here for e-commerce, we empirically define keywords as lemmatized, non-stopping nouns, verbs and adjectives appearing in \textit{questions}, according to our observations on specificity (\S \ref{sec:specific}). Note that keywords are different from \textit{answers}, and we don't assume the 
existence of an answer in our approach. We extract ground truth keywords and a keyword dictionary 
$Z$ of size $C$ from the CQs in the training set using this definition. 

With keywords introduced, the marginal likelihood of a question are 
decomposed as:

\begin{equation}
  \begin{split}
    p(\mathbf{y}|\mathbf{x}) &= \sum_{\mathbf{z} \subseteq Z}p(\mathbf{y},\mathbf{z}|\mathbf{x}) \\ 
    &= \sum_{\mathbf{z} \subseteq Z}p(\mathbf{y}|\mathbf{x},\mathbf{z})p(\mathbf{z}|\mathbf{x})
  \end{split}
  \label{equ:decompose}
\end{equation}
where $p(\mathbf{z}|\mathbf{x})$ corresponds to the keyword prediction part, and $p(\mathbf{y}|\mathbf{x},\mathbf{z})$ refers to the keyword conditioned generation. The range of $\mathbf{z} \subseteq Z$ is very large, so in practice, we sample portions of them to get an approximation, as will be discussed later (\S \ref{sec:selection}).

\subsection{Specificity}
\label{sec:specific}
In this work, the specificity of a question is determined by the size of 
its applicable range. Question that can only be raised against one particular context is 
considered more specific than universal questions. 
First, \textit{relevance} of the question is the basic
requirement of specificity.  Traditional MLE training may generate generic but not 
relevant question for higher likelihood. We conjecture that the 
additional task of keyword prediction will help focus on relevant aspects. 
Moreover, by observation, we discover that \textit{specificity} of e-commerce questions can be further promoted by: 
\begin{enumerate}
  \item Focusing on certain aspects, like the type, brand and attributes.
  \item Mentioning components of a product, e.g. blade of a food grinder.
  \item Describing a using experience specific to the product, such as cooking rice in a cooker.
\end{enumerate}

We hypothesize that many of them can be captured by keywords, with nouns and adjectives covering aspects and components, and verbs constituting the using experience.

\subsection{Diversity}
\label{sec:diversity}

Diverse CQGen requires a group of questions to be generated about the same context, to cover various information needs as well as improve the robustness to problematic generations. This setting differs from some previous 
literature~\citep{wang2018learning, rao2019answer}, where they generate 
only one response at a time, and \textit{Diversity} is used to measure 
the expected variety among 
\textit{all generated response}. We call it \textit{global diversity}. 
Our setting is referred to as \textit{local diversity}, measuring 
the diversity \textit{within one usage}. This is also adopted by another 
line of literatures \citep{vijayakumar2018diverse, shen2019mixture}. 
If not specified, we mean \textit{local diversity} by using \textit{diversity}. 
Global diversity is also desired, as it increases the likelihood of the questions to be specific to various contexts. 

To meet the diversity requirement as well as to promote specificity, we propose KPCNet below.

\section{Model Description}
\label{sec:KPCNet}

\begin{figure}[htbp]
  \centering
  \includegraphics[width=1\linewidth]{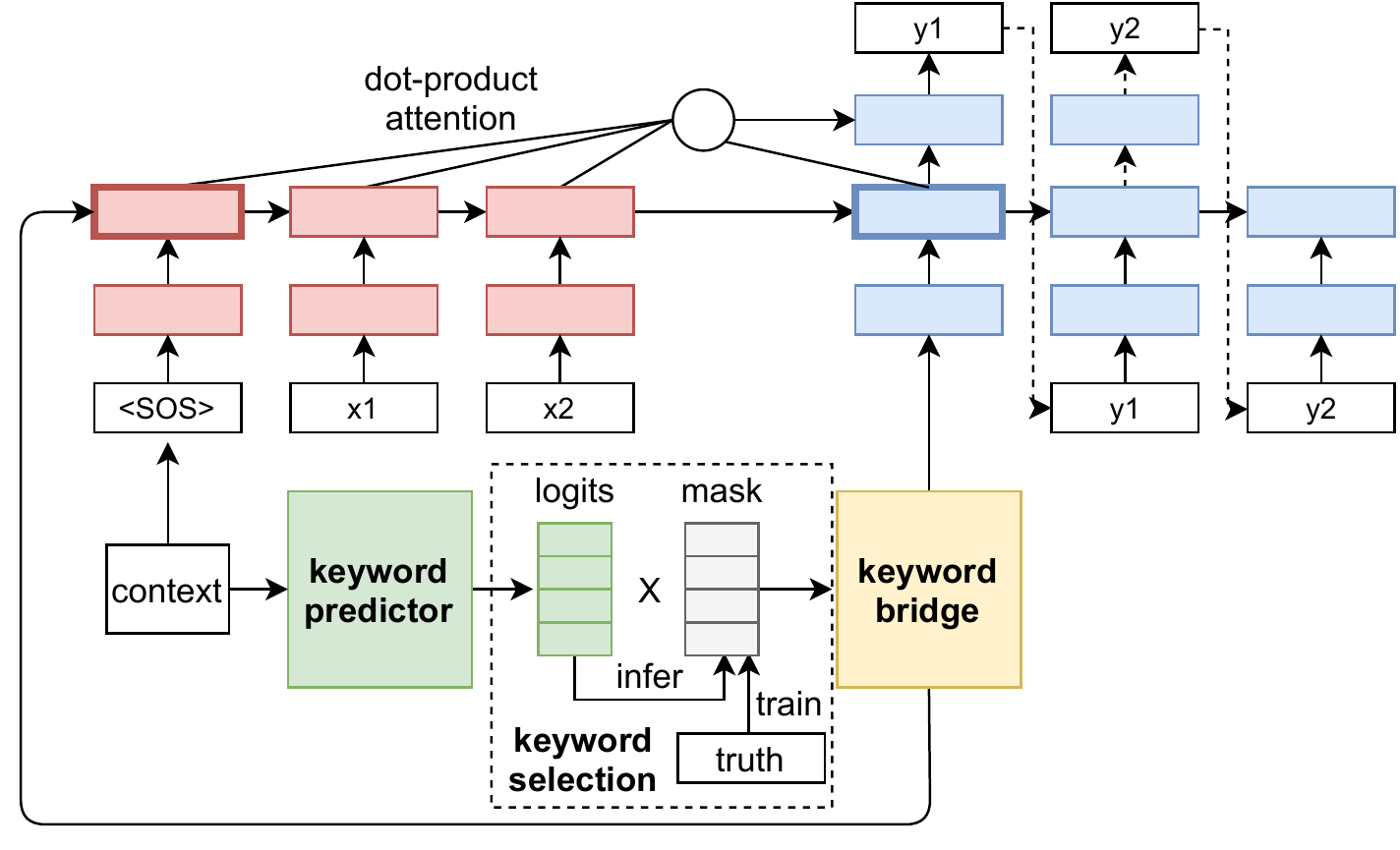}
  \caption{Illustration of KPCNet.}
  \label{fig:pipeline}
  \end{figure}

  In Equation \ref{equ:decompose}, $p(\mathbf{z}|\mathbf{x})$ corresponds to the keyword prediction part, and $p(\mathbf{y}|\mathbf{x},\mathbf{z})$ refers to the keyword conditioned generation. Our model is thus divided into 2 logical parts. The whole model pipeline is illustrated in Figure \ref{fig:pipeline}.  

\subsection{Keyword Prediction}
\label{sec:prediction}

For the \textit{Keyword Predictor}, we assume the probability of each keyword $z$ are independent from each other given context $\mathbf{x}$, i.e. $p(\mathbf{z}|\mathbf{x})=\Pi_{z \in \mathbf{z}}p(z|\mathbf{x})$, to simplify the modeling. We parameterize $p(z|\mathbf{x})$ with TextCNN \citep{kim-2014-convolutional}. The training loss is binary cross entropy over each keyword:

\begin{equation}
  L_{pred} = -\frac{1}{N}\sum_{n=1}^{N}\sum_{c=1}^{C}z^t_{n,c}log(p_{n,c})
  \label{equ:pred}
\end{equation}
Here we use $z^t_{n,c}$ as a binary indicator that shows if $c_{th}$ keyword in keyword dictionary $Z$ is among the ground truth keywords of the $n_{th}$ sample, and $p_{n,c}$ is the predicted probability for it.

\subsection{Keyword Conditioned Generation}
\label{sec:cond_gen}

The main structure of our generator is based on a standard sequence-to-sequence model \citep{luong2015effective}. We will focus on our specific design to condition the generation on keywords. 

\paragraph{Keyword Selection} We take the unnormalized keyword logits $\hat{p} \in \mathbb{R}^C$ from the keyword predictor, and then we select a conditioning keyword set $\mathbf{z}^s$ to mask out irrelevant dimensions to get a masked logits $\tilde{p} = [\hat{p}_1 z^s_1, \hat{p}_2 z^s_2, ..., \hat{p}_C z^s_C]$. This procedure allows us to control the generation with the selected keywords. Specific methods for this part will be discussed in \S \ref{sec:selection}.

\paragraph{Keyword Bridge} After getting the masked logits $\tilde{p}$, we pass them through a dropout layer, and then transform them to another distributed representation using a Multi-Layer Perceptron (MLP). They are then transformed into encoder features and decoder features with 2 MLPs respectively. The encoder feature will replace the hidden state of the first encoder step as memory to guide the generation via attention. The decoder feature will be fed as the input word embedding of the first decoder step to influence the generation.

\subsection{Keyword Selection}
\label{sec:selection}

At training, the ground truth keywords set $\mathbf{z}^t$ is selected as $\mathbf{z}^s$, and the training objective is to maximize the log-likelihood of all questions given context $\mathbf{x}$ and keywords $\mathbf{z}^t$. This equals to minimize: 

\begin{equation}
  L_{mle} = -\frac{1}{N}\sum^N_{n=1}log(p(\mathbf{y}_n|\mathbf{x}_n,\mathbf{z}^t_n))
  \label{equ:mle}
\end{equation}

At inference, we select $\mathbf{z}^s$ from keyword predictor's predicted distribution as condition for generation. This process was done once at a time, and can be done several times to fully explore the diversity in $p(\mathbf{y}|\mathbf{x})$ (Equation \ref{equ:decompose}) with different keyword sets. We come up with 3 methods for keyword selection:

\paragraph{Threshold} We select all keywords whose predicted probability are above a threshold $\alpha$ as $\mathbf{z}^s$. If not specified, this is the default selection method at inference.

\paragraph{Sampling} The threshold selection approach is deterministic and thus limited to one conditioning keyword set. We may encourage more diverse generation via diversifying the keyword set. An intuitive solution is to introduce randomness. Inspired by the Top-K \citep{fan-etal-2018-hierarchical, radford2019language} and Top-p (nucleus) sampling \citep{holtzman2019curious}, we also adopted a similar approach, sampling $k$ keywords from softmax-normalized prediction distribution after Top-K, Top-p filtering.

\paragraph{Clustering} Both the threshold and sampling selection strategies 
run the risk of putting semantically uncoherent keywords together, 
which is the drawback of the independence assumption used by keyword predictor. 
For example, if \textit{``voltage, machine, long, waffle''} are selected as 
the keywords for the waffle maker in Figure \ref{fig:WA_UI}, 
we may generate an illogical question ``what are the \textit{voltage} of 
the \textit{waffle}''. To get more coherent keyword sets, 
we explore the use of clustering technique. For the above example, 
the keywords can form 2 semantic groups, which lead to 
``What is the \textit{voltage} of the \textit{machine}'' and 
``How \textit{long} does it take to cook \textit{waffle}'', respectively. 

In practice, we first mine a keyword co-occurrence graph from the training set. We then take the Top-K likely keywords, and run Spectral clustering \citep{Shi00normalizedcuts} on the induced subgraph of them. The resulting $g$ disjoint groups are then used as generation conditions respectively.

\subsection{Keyword Controllability Probing}
\label{sec:probing}

One potential benefit that KPCNet brings is the controllability over 
generation by providing different conditioning keywords. To probe into this, 
we propose 2 approaches to operate on the keywords besides the 3 keywords 
selection methods. The operations are designed with hypotheses that 
will be tested with experiments.

\begin{table}[th]
  \small
  \centering
\begin{tabular}{l|l}
\hline
Product & \makecell[l]{iliving organic buckwheat pillow with authentic \\ japanese pillow cover, 14 by 20-inch, green } \\
\hline
KPCNet & \makecell[l]{what is the \textbf{size} of this \textbf{pillow} case? \\ (size, cover, pillow, wash, zipper)} \\
+Filter        & \makecell[l]{does this \textbf{pillow} have a \textbf{zipper}? \\ (cover, pillow, wash, zipper)} \\
\hline
\end{tabular}
\caption{\label{tab:kwd-filter} Example on the effect of keyword filtering. Predicted keywords for a question are shown in the parentheses below. ``size'' was filtered as it has already been covered in product description.}
\end{table}

\subsubsection{Keyword Filtering} \label{para:filter} Asking only about
things \textbf{not} in the context is the basic requirement of CQGen. 
However, none of existing methods in the literature 
have specific solution for this. In preliminary experiments of KPCNet, 
we found that some of the repetitive cases came with repetitive keywords.
Therefore, we conjecture that we may alleviate the problem by filtering 
out such repetitive keywords. Table \ref{tab:kwd-filter} provided a 
concrete example. This would be especially useful for iterative generation, 
as we will explicitly exclude repeating keywords if user triggers CQGen 
for the second time with some information vacancy already filled.

Here we use a simple matching method for keyword filtering. 
We first select a set of keywords that tends to lead to repetition. 
Then for each keyword in the set, we maintain a blacklist of words or 
patterns so that we filter the keyword if the pattern is matched. 
For example, we would filter words like ``height''/``width'' from the predicted keywords, if we can match ``height''/``width'' in the context. 
This process is currently done manually, so it doesn't scale. 
However, we find that a small set of frequent keywords is already 
enough to cover a relatively large number of repetitive cases and 
demonstrate the effect of this approach, as will be shown in \S \ref{sec:experiments}. 
We leave automatic repeating keyword detection and filtering for future works.

\subsubsection{External Knowledge} \label{sec:knowledge}
It is a common practice for e-commerce platforms to build knowledge graph 
to manage their products~\citep{dong2018challenges, luo2020alicoco}. 
As a result, products are attached to highly related tags, concepts, 
or keywords in our terms. Since the keywords used here is just a simple kind of knowledge, we believe that such richer external knowledge may further improve the generation by directly providing high-quality keywords, or helping the keyword prediction. Nevertheless, since we don't have access to such knowledge, we simulate such scenario where we have higher quality keywords by directly feeding ground truth keywords to the model[KPCNet(truth)]. This establishes an upper bound to what extent can KPCNet be improved with knowledge.

\subsection{Deduplication Postprocessing}
\label{sec:deduplicate}
All algorithms will more or less produce semantically similar questions 
in their initial generation group. Therefore, we will first generate 
more candidates than needed (say, produce 6 questions for 3 displaying slots), 
so that at least certain level of diversity can be guaranteed for 
the initial group. We then apply a simple, model-agnostic heuristic for 
deduplicating question selection. We first add the top generation into 
the current group, then we will iterative through the remaining questions. 
If the question's Jaccard similarity with any currently selected question 
is below 0.5, it will be added into the current group, otherwise 
it will be discarded.

\section{Experiments}
\label{sec:experiments}
In this section we try to answer the following research questions: 
\begin{enumerate}
  \item Can KPCNet generate more specific CQs than previous baselines? 
  \item To what extent can we control the generation of KPCNet by operating on the keywords with methods like keyword selection and filtering (\S \ref{sec:selection}, \S \ref{sec:probing}) ?
  \item How well can our proposed keyword selection methods promote local diversity, compared to existing diverse generation approaches?
\end{enumerate}

\subsection{Evaluation metrics}
Most previous works on question 
generation~\citep{jain2017creativity, hu2018aspect, rao2019answer} 
adopts \textit{Individual-level} evaluation protocol, where only the best generated question of a group is evaluated (thus also named \textit{Oracle} metrics). Specially, for proper evaluation of the novel \textit{Diverse CQGen} task, we need to evaluate the overall quality and diversity of CQ groups. We refer to this as \textit{Group-level} evaluation. We adopt automatic metrics as well as human judgements on both level. 

\subsubsection{Automatic Metrics}
We use \textbf{Distinct-3} (DIVERSITY), 
\textbf{BLEU}
\footnote{\url{https://github.com/moses-smt/mosesdecoder/blob/master/scripts/generic/multi-bleu.perl}}
\citep{papineni2002bleu} and 
\textbf{METEOR} \citep{banerjee2005meteor} for individual-level automatic evaluation.
For group-level evaluation, we adopt the evaluation protocol proposed by \citet{shen2019mixture} for diverse machine translation, and use \textbf{Pairwise-BLEU} and \textbf{Avg BLEU} as the evaluation metric. We report them in percentage.
\subsubsection{Human Judgements}
For individual-level human judgements, we show every annotator one context and one generated question for each system (including reference). The system name is invisible to the annotator and the order is randomly shuffled. The selected candidate is the one that achieved the highest BLEU in the generation group. 
We ask human to judge the \textbf{Grammaticality(G), Relevance(R), Seeking New Information(N) and Specificity(S)} of the questions. Also, noting that the system generations are also prone to make logical errors like improper repetition (``does the lid have a lid ?'') or asking for relevant but not exactly the correct object (asking ``what is the thickness of the bed ?'' for a mattress), we further judge the \textbf{Logicality(L)} of the candidate. Futher descriptions of these metrics can be found in Appendix B.

For group-level human judgements, we run the deduplication procedure (\S \ref{sec:deduplicate}) to get 3 top questions for each system. And annotators are showed one context and the 3 selected questions for each group. The groups are also anonymized and shuffled.

For each question in a group, we score the same metrics as those for individual-level judgements. To evaluate the valid variety of each group produced by local generation diversity, we introduced an additional and important group-specific metric: \textbf{\#Useful}. This is the number of useful questions after excluding problematic (ungrammatical, irrelevant, illogical, etc.) and semantically equivalent questions within a group. And we further calculate \textbf{\#Redundant} as (the number of unproblematic questions - \textbf{\#Useful}) to measure local redundancy.

Individual-level and group-level evaluation was conducted on the same set of 100 sample products for 8 systems and every group has 3 questions. They are distributed to 4 annotators so that each of the 2400 questions are annotated twice. We report inter-annotator agreement in Appendix B.

\subsection{Dataset}
We evaluate our model on the \texttt{Home \& Kitchen} category of the Amazon Dataset \citep{mcauley2015image,mcauley2016addressing} preprocessed by \citet{rao2019answer}. We apply extra preprocessing on the raw data to remove noises in dataset (see Appendix A). In this dataset, \textit{context} is the product title concatenated with the product description, and \textit{question} is the CQ asked by customers to the product. It consists of 19,119 training, 2,435 validation and 2,305 test examples (product descriptions), with 3 to 10 questions (average: 7) per description. The inherent diversity of questions in the dataset allows the proper evaluation of group-level generation diversity. We process another category, \texttt{Office}, in a similar way. \texttt{Office} is a much smaller dataset, consisting of 2,190 training, 285 validation and 256 test examples, with 3 to 10 questions (average: 6) per description. We will first analyze the results on \texttt{Home \& Kitchen} in detail, then briefly discuss the results on \texttt{Office}.

\subsection{Baselines}

For individual-level generation, we compare KPCNet with the following models: 
\paragraph{MLE} Vanilla seq2seq model trained on (context, question) pairs using maximum likelihood objective.  
\paragraph{hMup} A representative of the family of mixture models proposed 
by \citet{shen2019mixture}, which achieved a good balance of 
overall quality and diversity. 

Since we don't assume the availability of answers, we don't include traditional QGen methods and GAN-Utility \citep{rao2019answer} in the comparison.
For a fair comparison, we control the encoder and decoder for all the above methods to have a similar 2-layer GRU \citep{cho2014learning} or LSTM \citep{hochreiter1997long} architecture and close amount of parameters. 

\vspace{0.8em}

For group-level generation, we compare across 3 categories of diverse generation methods:
\paragraph{Decoding based} Classical beam search naturally produces different generation on each beam. Therefore, we evaluate the effect of beam search combined with MLE and KPCNet with threshold selection [KPCNet(beam)]. Recently, several decoding approaches \citep{ippolito2019comparison} are proposed to further promote diversity in generation, among which \textit{Diverse Beam Search}\citep{vijayakumar2018diverse} and \textit{Biased Sampling} like top-K, top-p sampling \citep{fan-etal-2018-hierarchical, holtzman2019curious} are representative methods. So we also evaluate KPCNet with them [KPCNet(divbeam), KPCNet(BSP)].
\paragraph{Model based} hMup is designed for diversity at the model level. It provides a discrete latent variable called \textit{expert} to control the generation. We thus take the top beam-searched candidate of each expert to form a generation group for evaluation.
\paragraph{Keywords based} This is dedicated to KPCNet. We evaluate the \textit{Sampling}[KPCNet(sample)] and \textit{Clustering}[KPCNet(cluster)] methods for keyword selection. We also estimate the potential of KPCNet with knowledge (\S \ref{sec:knowledge}) by providing the ground truth keyword set [KPCNet(truth)].

\vspace{0.8em}

All systems using beam search have a beam size of 6, we also set number of experts for hMup to 6, and we use beam size of 6 with 3 diverse groups for \textit{diverse beam search}. We select 2 keyword groups for KPCNet(sample) and KPCNet(cluster). To produce the final generation group for evaluation, outputs of all systems will go through the same deduplication postprocessing (\S \ref{sec:deduplicate}) to get 3 questions for each group.

\subsection{\texttt{Home \& Kitchen} Dataset Results}

\subsubsection{Individual-level Evaluation}

\begin{table}[htbp]
  \small
  \centering
  \begin{tabular}{l|ccccc}
  \hline
  {} & Distinct-3 & BLEU & METEOR \\
  \hline
  ref  &        69.34 &        - &    - \\
  \hline
  MLE &        7.77 &  \textbf{18.13} & 14.86 \\
  hMup &        11.11 &  17.76 &    15.40  \\
  KPCNet &        \textbf{15.30} &     17.77 &    \textbf{16.18}  \\
  \hline
  KPCNet(truth) &        37.38 &     23.63 &    19.38  \\
  \hline
  \end{tabular}
  \caption{\label{tab:ind-auto-eval} Individual-level automatic evaluation results on \texttt{Home \& Kitchen} dataset.}
\end{table}

Table \ref{tab:ind-auto-eval} shows the automatic evaluation results. KPCNet and hMup outperform MLE in METEOR but not in BLEU. We claim that it is due to the shorter and the safer generation of MLE, which is naturally favored by precision-based BLEU but not F-based METEOR. The average generation length is 5.957 for MLE, 8.231 for hMup, and 7.263 for KPCNet. KPCNet significantly outperform all the other baselines in Distinct-3 and METEOR, showing that KPCNet potentially promote higher global diversity and generation quality. We note that KPCNet(truth) has a great advantage over KPCNet, indicating the controllability of keywords and the potential of KPCNet to be further strengthened by improving the conditioned keyword set with other helpers like external knowledge (\S \ref{sec:knowledge}).

\begin{table}[htbp]
  \small
  \centering
  \begin{tabular}{l|ccccc}
  \hline
  {} & G & R & L & N & S \\
  \hline
  ref  &        0.98 &        1.00 &    1.00 &     0.94 &     2.68 \\
  \hline
  MLE  &        \textbf{0.99} &     0.92 &    \textbf{0.98} &     \textbf{0.85} &     1.45 \\
  hMup &        \textbf{0.99} &     0.92 &    0.86 &     0.81 &     1.81 \\
  KPCNet &        \textbf{0.99} &     \textbf{0.99} &    0.95 &     0.80 &     1.81 \\
  KPCNet(filter) &        \textbf{0.99} &     \textbf{0.99} &    0.94 &     \textbf{0.85} &     \textbf{1.84} \\
  \hline
  \end{tabular}
  \caption{\label{tab:ind-human-eval} Individual-level human evaluation metrics on 100 sample products from \texttt{Home \& Kitchen}. G/R/L/N/S stand for Grammaticality, Relevance, Logicality, New Info and Specificity respectively.}
\end{table}

\begin{table*}[htbp]
  \centering
  \small
  \begin{tabular}{l|ccccccc}
    \hline
    {} & Relevant\tiny{[0-1]} & Logical\tiny{[0-1]} & New Info\tiny{[0-1]} & Specific\tiny{[0-4]} & \#Useful\tiny{[0-3]} & \#Redundant\tiny{[0-2]} & Avg Rank \\
    \hline
    ref          &    0.990 &   1.000 &    0.947 &    2.530 &   2.680 &      0.120 & - \\
    \hline
    MLE          &    0.907 &   \textbf{0.943} &    \textbf{0.863} &    1.457 &   1.550 &      0.590 & 3.667 \\
    hMup         &    0.900 &   0.793 &    0.833 &    1.727 &   1.530 &     \textbf{0.130} & 4.667 \\
    \hline
    KPCNet(-filter)  &    \underline{\textbf{0.987}} &   \underline{0.870} &    \underline{0.830} &    1.757 &   1.280 &      0.750 & 4.500 \\
    KPCNet(beam)    &    \underline{\textbf{0.987}} &   \underline{0.853} &    \underline{\textbf{0.863}} &    1.793 &   1.330 &      0.750 & 3.667 \\
    KPCNet(divbeam) &    \underline{0.963} &   0.780 &    \underline{0.860} &    1.760 &   1.480 &  \underline{0.310} & 4.167 \\
    KPCNet(sample)  &    \underline{0.963} &   \underline{0.837} &    \underline{0.850} &    \underline{\textbf{1.890}} &   1.500 &      0.450 & 3.500 \\
    KPCNet(cluster)  &    \underline{0.963} &   \underline{0.863} &    \underline{0.823} &    \underline{1.877} &   \underline{\textbf{1.760}} &      \underline{0.190} & \textbf{3.000} \\
    \hline
    \end{tabular}
  \caption{\label{tab:group-human-eval} Group-level human evaluation results on 100 sample products (300 questions each system) from \texttt{Home \& Kitchen}. Grammaticality is omitted as the results are similar to Table \ref{tab:ind-human-eval} where all systems performs well. \textit{Avg Rank} is the average ranking among all 7 methods across the 6 metrics. We perform hypothesis test among KPCNet variants, and the difference between underlined and non-underlined numbers at each column is statistically significant with $p \leq 0.05$.}
  \end{table*}

Table \ref{tab:ind-human-eval} shows the individual-level human evaluation results. We can see that all systems perform well in \textit{Grammaticality}, 
KPCNet significantly outperforms other systems in \textit{Relevance} and 
achieved the best \textit{Specificity}, while performs slightly worse in \textit{Logicality}. The superior \textit{Relevance} score validates our hypothesis 
that independently trained keyword predictor help focus on relevant keywords 
instead of irrelevant but generic words preferred by MLE 
(\S \ref{sec:specific}). KPCNet(filter) gets a much higher 
\textit{New Info} at the cost of only slight drop in \textit{Logicality}. 
It shows that the Keyword Filtering step (\S \ref{para:filter}) can truly 
utilize the controllability of keywords to help avoid repetition on the 
basis of KPCNet. Therefore, we by default incorporate the step with 
all the KPCNet variants in the next group-level evaluation stage while 
keeping the vanilla KPCNet for comparison as KPCNet(-filter).

\begin{figure}[htbp]
  \centering
  \includegraphics[width=\linewidth]{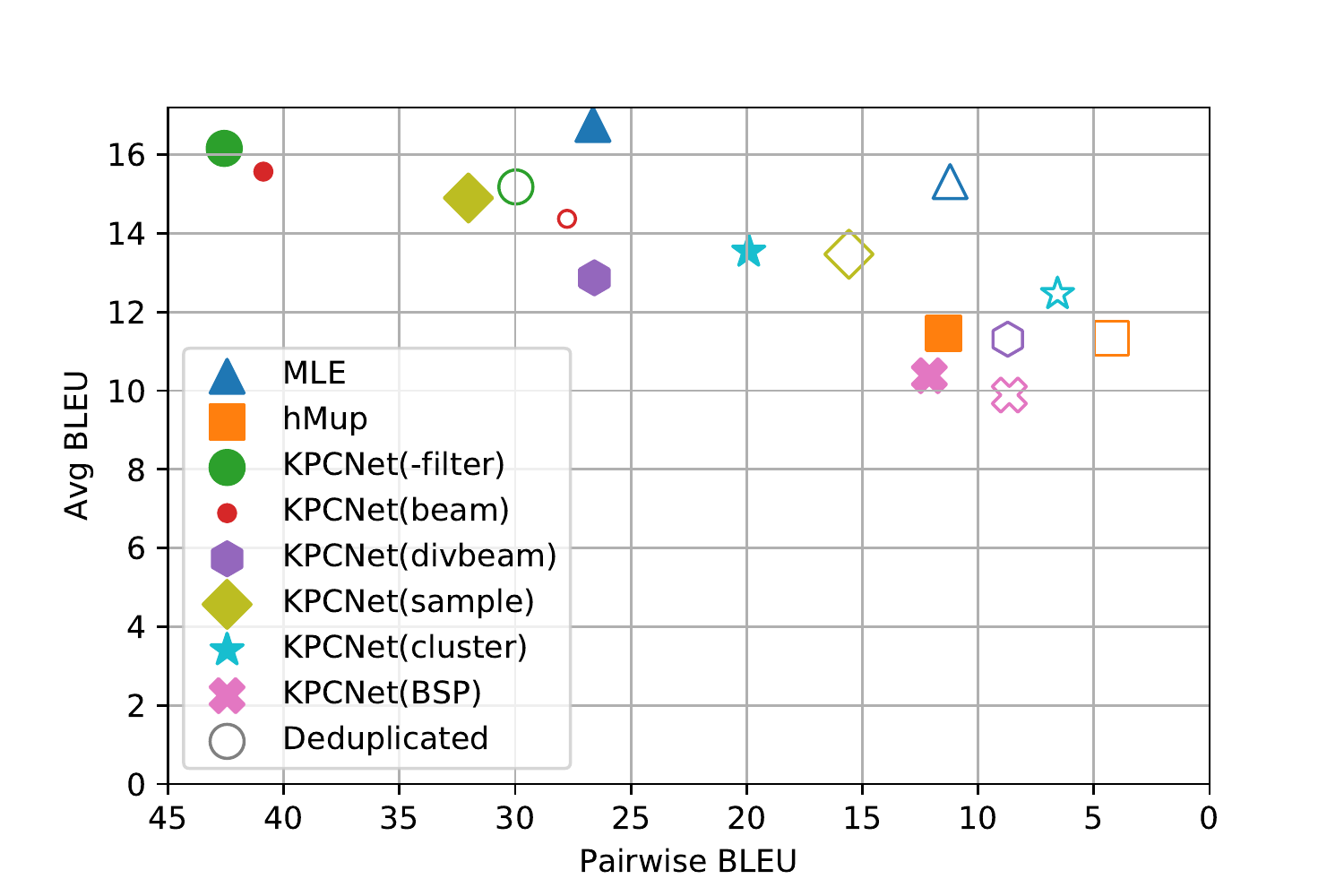}
  \caption{Group-level Automatic metrics on the whole test set of \texttt{Home \& Kitchen}. The lower Pairwise BLEU, the more diverse the generated group. Solid markers are the results for the top 3 candidates in the original group, while hollow markers measures the remaining 3 after deduplication. Points located near top-right are preferred as they achieve a good tradeoff between the 2 metrics.}
  \label{fig:group-filter}
  \end{figure}

\subsubsection{Group-level Evaluation}

The group-level automatic evaluation metrics before and after deduplication for each system are shown in Figure \ref{fig:group-filter}. Original results are shown in solid markers. KPCNet(BSP) has the poorest Avg BLEU and we found the results very likely to be ungrammatical and illogical, and we thus omit it in the following evaluation. hMup has the highest local diversity while has the second poorest Avg BLEU. MLE has moderate level of local diversity and the highest Avg BLEU, and we found that Keyword Filtering slightly harmed Avg BLEU, which is against our intuition. But we later found Avg BLEU doesn't correlate well with most human judgements (discuss later). Several diversity-promoting variants of KPCNet improved local diversity at the cost of Avg BLEU, among which KPCNet(cluster) achieved a best tradeoff between the two. Comparing the original and deduplicated results (hollow markers), we can see that our simple heuristic can effectively eliminate redundancy at the cost of slight degradation of Avg BLEU, as only nearly identical hypotheses with high BLEU are excluded. 

Group-level human evaluation results are shown in Table \ref{tab:group-human-eval}. We can see that all KPCNet variants clearly outperform baselines in \textit{Relevant} and \textit{Specific} while have a competitive performance in \textit{New info}. MLE rated best for \textit{Logical} for its conservative generations (low \textit{Specific}), and the questions tend to overlap with each other, as is reflected in high \textit{\#Redundant}. KPCNet(beam) has a even higher redundancy since its searching space is further limited by the conditioned keyword set. Diverse generation variants can help overcome this drawback. Especially, KPCNet(cluster) achieved the best \textit{\#Useful}, \textit{Avg Rank}, and its performance on all metrics is among the best of KPCNet variants. This shows that the semantically-coherent keyword sets produced by clustering can effectively improve the generation diversity and quality of KPCNet. 

We also study the system-level Pearson correlation between the automatic metrics and human judgements. Pairwise-BLEU has a correlation  of 0.915 with \textit{\#Redundant} ($p<0.01$), -0.835 with \textit{\#Useful} ($p<0.05$). Avg BLEU is shown only correlates well with \textit{Logical} (correlation: 0.849, $p<0.05$). This result validates the use of Pairwise-BLEU as an automatic proxy metric for local diversity.

\subsubsection{Case Study}

\begin{table*}[htbp]
  \centering
  \begin{tabular}{c|lcc}
      \hline
      product & \makecell[l]{homelegance 2588s accent dining chair, blue grey, set of 2} & {} & {} \\
      \hline
      \makecell[c]{system \\ (\#Useful)} & generation group & specific & problem \\
      \hline
      \makecell[c]{ref \\ (3)} & \makecell[l]{can any of the recent reviewers confirm the seat height ? \\ i see the question was posted in april ... \\ would u please send me the box dimensions ( when buy in \\ a set of 2 ) and the weight ? \\ can someone please tell me the depth of the chair seat \\ from the end of the curved back to the end of the seat ? } & \makecell[c]{2 \\ \\ 3 \\ \\ 3 \\ \\} & {} \\
      \hline
      \makecell[c]{MLE \\ (1)} & \makecell[l]{what is the seat height ? \\ what are the dimensions of the chair ? \\ what are the dimensions ?} & \makecell[c]{2 \\ 2 \\ 1} & {} \\
      \hline
      \makecell[c]{hMup \\ (1)} & \makecell[l]{what is the weight limit for the chair ? \\ i have a table that is a [UNK]. will this chair be able \\ to fit on a table ? \\ is this a set of 2 chairs or just one ?} & \makecell[c]{2 \\ 2 \\ \\ 2} & \makecell[c]{ \\ illogical \\ \\ repetitive} \\
      \hline
      \makecell[c]{KPCNet \\ (2)} & \makecell[l]{what is the \textbf{color} of the \textbf{chair} ? \\ what are the \textbf{dimensions} of the \textbf{seat} ? \\ what is the \textbf{weight} limit ? } & \makecell[c]{2 \\ 2 \\ 2} & \makecell[c]{ repetitive \\ \\  \\ } \\
      \hline
      \end{tabular}
      \caption{\label{table:quality} Example generation group and the human judgements for each system. Here we use KPCNet to stand for KPCNet(cluster) for brevity, and the appeared keywords of KPCNet are in bold. }
\end{table*}

\begin{table}[htbp]
  \small
  \centering
  \begin{tabular}{l|l}
  \hline
  Product & \makecell[l]{Novaform memory foam comfort curve pillow} \\
  \hline
  \makecell[l]{KPCNet \\ (cluster)} & \makecell[l]{is this a \textbf{firm} \textbf{pillow}? (pillow, foam, sleep, firm) \\ is this pillow good for \textbf{stomach sleepers}? \\ (stomach, sleeper)} \\
  \hline
  Product & \makecell[l]{full-sized headboard in solid wood} \\
  \hline
  \makecell[l]{KPCNet \\ (cluster)} & \makecell[l]{what is the height of this \textbf{headboard} ? \\ (bed frame headboard) \\ does it have a \textbf{box spring} ? (mattress box spring)} \\
  \hline
  \end{tabular}
  \caption{\label{tab:kwd-cluster} Example generation groups for KPCNet(cluster). Keywords in the parentheses.}
  \end{table}

Table \ref{tab:kwd-cluster} provides 2 example generation groups of KPCNet(cluster). For each group, the 6 predicted keywords captured specific aspects of the product. Then they are divided into 2 coherent groups (as they formed natural phrases such as ``firm pillow'' and ``stomach sleeper'') by clustering. Finally, the different conditioned keyword sets are reflected in the generation. In the first case, specific and diverse generations are successfully produced with precisely predicted keywords. We can see that the separation of keywords as controlling factors allows the novel use of classical clustering technique to help generate high-quality question groups by first producing coherent keyword sets. There are also bad cases like the second question in another group. The possible reason is that keyword predictor produced related but unsuitable keywords ``box spring'', which can be asked for a whole bed but not for headboard alone. This shows that predictor is the performance bottleneck of KPCNet.

We provide a group-level evaluation example in Table \ref{table:quality}. We can see that the diversity of MLE is very limited (it gets \textit{\#Useful} of only 1, though all 3 questions are valid, and thus \textit{\#Redundant} is 2), and it produces highly generic question. The generations are more diverse for hMup. However, we find that a certain expert of hMup has a style of long and illogical generation, like the second one demonstrated here. (It's abnormal to put chairs \textit{on} a table, and the text is not coherent as it doesn't use a pronoun in the second sentence.) This may attribute to its focus on \textit{style} instead of aspects of the products, as it is originally proposed for translation of diverse styles. This significantly harms hMup's group-level performance (Table 4) compared to its best single model (Table 3). KPCNet(cluster) produces a diverse and specific generation, and we can clearly see the effect of keyword in its generation. 

\subsection{\texttt{Office} Dataset Results}

For brevity, we only show the individual-level automatic evaluation and group-level human judgement results. All the experimental settings are the same with the previous experiments, except that we apply no keyword filtering here. 

\begin{table}[htbp]
  \centering
  \small
  \begin{tabular}{l|ccccc}
  \hline
  {} & Distinct-3 & BLEU & METEOR \\
  \hline
  ref  &        75.54 &        - &    - \\
  \hline
  MLE &        20.33 &  \textbf{14.73} & 13.81 \\
  hMup &        15.31 &  10.45 &    12.52  \\
  KPCNet &        \textbf{30.99} &     13.84 &    \textbf{15.29}  \\
  \hline
  \end{tabular}
  \caption{\label{tab:ind-auto-eval-office} Individual-level automatic evaluation results on the \texttt{Office} dataset.}
\end{table}

\begin{table*}[htbp]
\centering
\small
\begin{tabular}{l|ccccccc}
\hline
{} & Grammatical\tiny{[0-1]} & Relevant\tiny{[0-1]} & Logical\tiny{[0-1]} & New Info\tiny{[0-1]} & Specific\tiny{[0-4]} & \#Useful\tiny{[0-3]} & \#Redundant\tiny{[0-2]} \\
\hline
ref         &       0.993 &    0.997 &   0.993 &    0.933 &    2.713 &   2.420 &      0.330 \\
\hline
MLE         &       0.970 &    0.843 &   \textbf{0.883} &    0.797 &    1.470 &   1.070 &      0.420 \\
KPCNet &       \textbf{0.993} &    \textbf{0.940} &   0.817 &    \textbf{0.803} &    \textbf{1.903} &   \textbf{1.470} &      \textbf{0.190} \\
\hline
\end{tabular}
\caption{\label{tab:group-human-eval-office} Group-level human judgments on 100 samples from the \texttt{Office} dataset. KPCNet here uses keyword clustering.}
\end{table*}

Table \ref{tab:ind-auto-eval-office} shows that KPCNet still outperforms MLE in Distinct-3 and METEOR, while falls behind at BLEU. Both the automatic metrics and our manual check indicate that hMup fails to give comparable results for 
the small dataset, so we exclude it in group-level evaluation.

Table \ref{tab:group-human-eval-office} shows that the performance of both models degraded here possibly due to the smaller data size. However, the observation is similar. KPCNet(cluster) outperforms MLE in most metrics especially at \textit{Relevant}, \textit{Specific} and \textit{\#Useful} despite a weakness at \textit{Logical}. This shows that KPCNet(cluster) can consistently improve the diversity and specificity of the generation.

\section{Related Work}
\paragraph{Clarification Question Generation} The concept of CQ can be naturally raised in a dialogue system where the speech recognition results tend to be erroneous so that we raise CQs for sanity check \citep{stoyanchev2014towards}, or the intents for a task is incomplete or ambiguous in a first short utterance and further CQs are needed to fill in the slots \citep{dhole2020resolving}. The concept is then extended to IR to clarify ambiguous queries \citep{aliannejadi2019asking}, and has been successfully put into practice \citep{zamani2020generating}. Other application areas including KBQA \citep{xu2019asking} and open-domain dialogue systems \citep{aliannejadi2020convai3}. CQGen can also be applied to help refine posts on websites like StackExchange \citep{Kumar_2020} and Amazon \citep{rao2019answer}. In this context, our work closely follows the research line of \citep{rao2018learning, rao2019answer, cao2019controlling}. \citet{rao2018learning} first adopted a retrieval-then-rank approach. They \citep{rao2019answer} then proposed a generation approach to train the model to maximize the utility of the hypothetical answer for the questions with GAN, to better promote specificity. \citet{cao2019controlling} propose to control the specificity by training on data with explicit indicator of specificity, but it requires additional specificity annotation. Towards the similar specificity goal, we adopted a different keyword-based approach. They also assume generating one question per context, which we claim is not sufficient to cover various possible information needs, and thus propose the task of the diverse CQGen.

\paragraph{Diverse Generation} The demand for diverse generation exists in many other fields~\cite{vijayakumar2018diverse, LiangZ18code, shen2019mixture}, and we've drawn inspirations from these literatures. For image captioning, we may use multiple descriptions for different focusing points of a scene. \textit{Diverse Beam Search} \citep{vijayakumar2018diverse} was proposed to broaden the searching space to catch such diversity by dividing groups in decoding and imposing repetition penalty between them. For machine translation, a context can be translated with different styles. \citet{shen2019mixture} thus proposed \textit{Mixture of Expert} models including hMup to reflect various styles with a discrete latent variable (\textit{expert}). And here for CQGen, diversity is required to cover various potentially missing aspects, so we come up with the idea to use keywords as a controlling variable like \textit{expert} to promote diversity.

\section{Conclusion}
To tackle the problem of missing information in product descriptions on e-commerce websites, we propose the task of Diverse CQGen to request for various unstated aspects in the writing with a group of semantically different questions. We then propose KPCNet to deal with this novel task as well as improve the specificity of the questions with the prior knowledge on user needs in the form of keywords. Human judgements showed that KPCNet is able to generate more specific questions and promote better group-level diversity. Oracle tests with ground truth keywords provided in keyword selection showed strong performance, indicating the great potential to be exploited from improving keyword prediction possibly with external knowledge. Future works may include utilizing richer external knowledge to improve the keyword prediction, and solutions for the occasionally illogical generations. We also believe that our approach can be applied to other scenarios with slight domain-specific modifications on the utilized knowledge.

\begin{acks}
This work was partly supported by the
SJTU-CMBCC Joint Research Scheme and SJTU Medicine-Engineering
Cross-disciplinary Research Scheme.
\end{acks}

\bibliographystyle{ACM-Reference-Format}
\bibliography{paper}

\appendix

\section{Experimental Details}
\label{sec:detail}
\subsection{Data Cleaning}
The following steps are enforced to remove noises as well as remove unhelpful parts for the CQGen task in the original data:

\paragraph{Fixing Unescaped HTML characters} We noticed that there are unescaped HTML special characters in both context and the question. (e.g. ``does it slice like zucchini \textbf{\& amp ;} cucumbers?'' is changed to ``does it slice like zucchini \textbf{\&} cucumbers?'')

\paragraph{Remove non-question parts} Sometimes there are declarative sentences following the question, which is not the focus of our task. We thus removed them. (e.g. where is this product made ? \textit{i contacted customer service and the representative was uninformed and could not offer any information .})

\paragraph{Remove noise questions} Some questions contain the comparison between 2 specific entities, which is unlikely to be tackled by our model, so we dropped them. And some questions are too universal (``Does it ship to Canada?''). We consider them as noise and also dropped them.

Note that the data cleaning was only imposed on the training set and the validation set. We preserve exactly the same test set as \citet{rao2019answer} for fair comparison.

\subsection{Hyperparameters and other settings}
For all models, we set the max length of context to be 100, question to be 20. For all variants of KPCNet, we use 2-layer GRU \citep{cho2014learning} with 100 hidden units for both the encoder and decoder. We use a learning rate of 0.0003 to train at most 60 epochs. For MLE, the model structure and parameters are identical to KPCNet, and we follow the setting of \citet{rao2019answer}, using dropout=0.5, learning rate=0.0001 to train 100 epochs. To improve the generation quality, we block bigrams from appearing more than once, and also forbid 2 same words to appear within 3 steps. For sampling-based keyword selection, we sampled 3 keywords from top-$K$ top-$p$ filtered keywords distribution with $K=6, p=0.9$ for 2 times. For clustering-based keyword selection, we produce 2 clusters from the top 6 predicted keywords. For hMup, we use the implementation in fairseq\footnote{\url{https://github.com/pytorch/fairseq/blob/master/examples/translation_moe}}. The architecture is set to 2-layer LSTM \citep{hochreiter1997long} with 100 hidden units, and other settings are identical to KPCNet for fair comparison. The threshold $\alpha$ for the default keyword selection method of KPCNet is manually tuned within range [0.05, 0.1]. The dropout strength is shared among all components of KPCNet and is manually tuned within range [0.2, 0.5]. MLE and KPCNet is implemented in PyTorch. For all manually tuned hyperparameters, we fix all other hyperparameters and random search for value within given range that can achieve the best BLEU on our validation set. The models are trained on a Ubuntu 18.04.4 LTS server with one NVIDIA GeForce RTX 2080 Ti. 

For \texttt{Home \& Kitchen} dataset, all models are operated on 200D word embeddings borrowed from \citet{rao2019answer}, which are pretrained from in-domain data with Glove \citep{pennington2014glove} and are frozen during training, except for hMup, which uses unique embedding to distinguish between experts and thus the embeddings are trained from scratch. The selected threshold $\alpha$ is 0.07, after 3 trials, and the selected dropout is 0.3 after 4 trials.

For \texttt{Office} dataset, all models are operated on 200D word embeddings that we pretrained from in-domain data with Word2vec\citep{mikolov2013distributed} in gensim\footnote{\url{https://radimrehurek.com/gensim/models/word2vec.html}}, except for hMup. The selected threshold $\alpha$ is 0.07, after 3 trials, and the dropout is initially selected as 0.3 based on the result of \texttt{Home \& Kitchen}.

For hypothesis test in Table 4, we use \texttt{proportions\_ztest} of scipy for the first 3 columns whose range is binary, and \texttt{ttest\_rel} for the other 3 columns. The procedure we assign the underline are: First, we underline the best number at each column. Then we run hypothesis test against every other number. If the difference is not significant, we also underline it, otherwise we don't underline it. 

\section{Human Judgement Details}
\subsection{Metric Descriptions}
For human evaluation, we show each annotator a detailed annotation guideline with definitions and examples. Here we provide some brief explanations:
\begin{itemize}
  \item Grammaticality=0, if there is syntax error, or the generation result is not a question
  \item Relevance=0, if the problem is not related to the context
  \item Logicality=0, if there is clear nonsense within the question it self (does the lid have a lid ?), or the question is not suitable for the context (asking "how many bottles does it hold ?" for a bottle).
  \item Seeking New Information=0, if the question is asking for information already contained in the context, like asking color for a product titled "blue chair".
\end{itemize}

For Specificity, we ask “How specific is the question?”
and let annotators choose from:
\begin{itemize}
  \item 4: Specific pretty much only to this product (or same product from different manufacturer)
  \item 3: Specific to this and other very similar products
  \item 2: Generic enough to be applicable to many other products of this type
  \item 1: Generic enough to be applicable to any product under this category (H\&K or Office)
  \item 0: N/A (Not applicable) i.e. Question is ungrammatically, irrelevant or illogical
\end{itemize}

\subsection{Inter-annotator Agreement}
We report the inter-annotator agreement measured by Randolph's $\kappa$ \citep{randolph2005free} in Table \ref{tab:inter-annotator}. It can be seen that \textit{Grammatical} and \textit{Relevant} have high agreement as they are easy to judge. \textit{New Info} has lower agreement possibly because it is harder to decide. For the example in Table \ref{table:quality}, the question ``what is the color of the chair ?'' may have not been annotated as repetitive as the word ``color'' doesn't appear in the context, though it is actually covered by the specific value ``blue grey''. \textit{Logical} and \textit{Specific} have the lowest degree of agreement as they are more subjective criteria. According to the table suggested by \citet{landis1977measurement}, all the criteria achieved at least moderate agreement.

\begin{table}[htbp]
  \centering
  \begin{tabular}{l|ccccc}
  \hline
  Criteria & Agreement \\
  \hline
  Grammatical\tiny{[0-1]}  &  0.933 \\
  Relevant\tiny{[0-1]} &  0.853 \\
  Logical\tiny{[0-1]} &  0.659  \\
  New Info\tiny{[0-1]} &  0.701  \\
  Specific\tiny{[0-4]} &  0.546  \\
  \hline
  \end{tabular}
  \caption{\label{tab:inter-annotator} Inter-annotator Agreement measured by Randolph's $\kappa$ \citep{randolph2005free}}
\end{table}

\section{Ablation Test}
\label{sec:ablation}

Below we describe the ablation test to check the influence of the components and hyperparameters of the model. These tests are all conducted on the \texttt{Home \& Kitchen} dataset.

\subsection{Additional Metrics}

To evaluate the quality of our keyword predictor and keyword bridge, we propose these additional automatic metrics:
\paragraph{P@5} Since the number of keywords in ground truth questions are different across each sample. We take the top 5 keywords with the highest predicted probability as selected keyword set $\mathbf{z}^s$, and calculates precision@5 by:
\begin{equation}
  P@5 = \frac{|\mathbf{z}^s \cap \mathbf{z}^T|}{5}
\end{equation}
where $\mathbf{z}^T$ is the union of keywords extracted from all ground truth questions of a sample.

\paragraph{Response Rate} which is the proportion of conditioned keywords that appears in the corresponding generation, and we report the macro average on all the records. We use this to evaluate the controllability of the keyword conditions.

We also report the average generation length(\textbf{Length}) as it is related to almost all metrics proposed above, but neither long or short generation should be considered an indicator of good performance.

\begin{table}[htbp]
  \centering
\begin{tabular}{l|ccccc}
\hline
{} & Distinct-3 & BLEU & P@5 & Response & Length \\
\hline
KPCNet(C, S) & 15.30 & 17.77 & 0.47 & 0.40 & 7.26 \\
-C & 16.51 & 15.88 & 0.51 & 0.35 & 7.52 \\
-S, +E & 12.00 & 9.04 & 0.22 & 0.50 & 7.66 \\
+H & 29.97 & 12.85 & 0.47 & 0.66 & 9.17 \\
\hline
\end{tabular}
\caption{\label{table:ablation1} Ablation test results on \texttt{Home \& Kitchen} for data and keyword predictor at individual-level. The first line is final adopted setting.}
\end{table}

\begin{table}[htbp]
  \centering
  \begin{tabular}{l|ccccc}
  \hline
  {} & Distinct-3 & BLEU & Response & length \\
  \hline
  Dropout = 0.2 & 17.29 & 17.11 & 0.45 & 7.16 \\ 
  Dropout = 0.3 & 15.30 & 17.77 & 0.40 & 7.26 \\
  Dropout = 0.4 & 13.02 & 18.33 & 0.35 & 6.95 \\
  Dropout = 0.4, NE & 15.04 & 18.19 & 0.34 & 6.78 \\
  Dropout = 0.4, ND & 12.19 & 17.47 & 0.32 & 6.53 \\
  Dropout = 0.5 & 11.77 & 18.53 & 0.32 & 6.66 \\
  \hline
  \end{tabular}
  \caption{\label{table:ablation2} Ablation test results for keyword bridge at individual-level on \texttt{Home \& Kitchen}.}
  \end{table}

  \begin{table}[htbp]
    \centering
    \begin{tabular}{l|ccccc}
    \hline
    {} & G & R & L & N & S \\
    \hline
    KPCNet &        \textbf{0.99} &     \textbf{0.99} &    \textbf{0.95} &     0.80 &     1.81 \\
    KPCNet(filter) &        \textbf{0.99} &     \textbf{0.99} &    0.94 &     0.85 &     \textbf{1.84} \\
    \hline
    KPCNet &        0.98 &     0.97 &    0.88 &     0.84 &     1.77 \\
    KPCNet(filter) &        0.98 &     0.97 &    0.89 &     \textbf{0.88} &     1.80 \\
    \hline
    \end{tabular}
    \caption{\label{tab:ind-human-eval-2} Comparison between KPCNet with Dropout=0.3 (upper half) and Dropout=0.2 (lower half) with individual-level human judgements on 100 sample products from \texttt{Home \& Kitchen}. G/R/L/N/S stand for Grammaticality, Relevance, Logicality, New Info and Specificity respectively.}
    \end{table} 

\subsection{Ablation Factors}

These are many important factors and parameters in our model. So we divide the ablation test into 2 logical parts: one for keyword predictor (and the effect of data cleaning on it), and another for keyword bridge. 

The ablation factors for keyword predictor are as follows (abbreviated for readability):
\begin{itemize}
  \item \textbf{E}: End2end training of keyword predictor with other component. The training objective is a weighted sum of the 2 objectives (Equation 2 \& 3).
  \item \textbf{S}: Separate training, first train predictor, and then freeze its parameters to train other parts. 
  \item \textbf{H}: Hard label fed to bridge instead of masked soft logits. The label can be provided from ground truth in training and is decided with threshold filtering in inference. If this setting works well, we can then completely separate the parameters of predictor from other parts.
  \item \textbf{C}: Cleaned dataset.
\end{itemize}

The ablation factors for keyword bridge are:
\begin{itemize}
  \item \textbf{NE}: No encoder feature fed back to encoder
  \item \textbf{ND}: No decoder feature fed to decoder
  \item \textbf{Dropout}: We add a dropout layer for the unmasked keywords logits before it passes the latter transformation. Due to the nature of dropout, this part may help ease the noise introduced by the error of keyword predictor. And we study the effect of the strength of this layer.
\end{itemize}

\subsection{Results}

The ablation test result for data and keyword predictor at individual-level is shown in Table \ref{table:ablation1}. The setting for keyword bridge is fixed: dropout=0.3, both encoder and decoder feature are used. After data cleaning(C), \textit{P@5} dropped because of the reduction of the number of ground-truth keywords. The decreasing of \textit{Distinct-3} and \textit{Length} shows the effect of irrelevant part removing. The improvement on \textit{BLEU} and \textit{Response} indicates the overall benefits brought by the cleaning. End2end training(-S, +E) leads to significant performance degradation on all metrics except slight increase on \textit{Response}. The possible reason is that keyword prediction skews highly towards frequent keywords under this condition. Finally, feeding hard label instead of logits also produce worse result. We can see from the extremely high \textit{Response} and \textit{Length} that this setting suffers severely from over-generation of keywords: model generates illogical long questions to contain as much keywords as possible. We hypothesize that the soft logits can reflect subtle difference on the importance of each conditioned keyword and thus can lead to more robust performance. Moreover, we can achieve a \textit{P@5} of 0.628 with one group of group truth keywords, as compared to 0.472 of the current model, which shows a huge room for improvement of the keyword predictor.

  The ablation test result for keyword bridge at individual-level is shown in Table \ref{table:ablation2}. The setting for keyword predictor is fixed as KPCNet(C, S). We can clearly witness the trend that the higher dropout, the higher controllability keywords will have over generation (Response). As a result, the behavior of KPCNet will be more and more like MLE when dropout grows, with lower generation length, lower keyword response and higher BLEU. We speculate that the dropout imposed on the keywords logits to be masked forces the model to make prediction with incomplete keyword set. Therefore, proper level of dropout can make the model robust to the noise introduced by keyword predictor. Furthermore, the ablation of either encoder bridge or decoder bridge would harm BLEU, response and length, which proved the effect of KPCNet's double-bridge design to guide the generation via attention between the two sides.

We also conducted human evaluation for different value of dropout (Table \ref{tab:ind-human-eval-2}), and found that lower dropout trades logicality for new information. We selected Dropout=0.3 as the final setting for its good balance of all metrics.

\end{document}